# Logarithmic-Time Updates and Queries in Probabilistic Networks

**Arthur L. Delcher**
Computer Science Department
Loyola College in Maryland
Baltimore, MD
delcher@cs.jhu.edu

**Adam Grove**
NEC Research Institute
Princeton, NJ
grove@research.nj.nec.com

**Simon Kasif**
Department of Computer Science
Johns Hopkins University
Baltimore, MD
kasif@cs.jhu.edu

**Judea Pearl**
Department of Computer Science
University of California
Los Angeles, CA
pearl@lanai.cs.ucla.edu

## Abstract

In this paper we propose a dynamic data structure that supports efficient algorithms for updating and querying singly connected Bayesian networks (causal trees and polytrees). In the conventional algorithm, new evidence is absorbed in time $O(1)$ and queries are processed in time $O(N)$, where $N$ is the size of the network. We propose a practical algorithm which, after a preprocessing phase, allows us to answer queries in time $O(\log N)$ at the expense of $O(\log N)$ time per evidence absorption. The usefulness of sub-linear processing time manifests itself in applications requiring (near) real-time response over large probabilistic databases.

## 1 INTRODUCTION

Probabilistic (Bayesian) networks are an increasingly popular modeling technique, which we can expect to be adopted in many applications that formerly might have used ordinary (non-probabilistic) database or expert-systems techniques. Many new applications will require real-time or near real-time operation, and so the question of doing *extremely* fast reasoning in Bayesian networks is important.

We are interested in settings with a repeated evidence-update and query cycle. That is, from time to time the observed values of certain evidence nodes change, and an update process absorbs this evidence into the network. A query is a request for the marginal probability of some other node. In the general case, of course, we cannot hope to reason quickly since the query process is known to be NP-hard [Coo90]. Even approximation is known to be hard [DL93]. However, the class of *singly-connected* networks (causal trees and polytrees) is known to admit efficient algorithms. But although the standard algorithms are linear in the size of the network, even this may be too slow for some purposes. In this paper, we show how to update and query polytrees with complexity that is only logarithmic in the number of nodes in the network.

We note that measuring complexity only in terms of the size of the network, $N$, can overlook some important factors. Suppose that each variable in the network has domain size $k$ or less. For many purposes, $k$ can be considered constant. Nevertheless, some of the algorithms we consider have a slowdown which is some power of $k$, which can be become significant in practice unless $N$ is very large. Thus we will be careful to state this slowdown where it exists.

Section 2 considers the case of *causal* trees; *i.e.*, singly connected networks in which each node has at most one parent. The standard algorithm (see [Pea88]) must use $O(k^2 N)$ time for either updates or for retrieval, although one of these operations can be done in $O(1)$ time. As we discuss briefly in Section 2.1, there is also a straightforward variant on this algorithm that takes $O(k^2 D)$ time for both queries and updates, where $D$ is the height of the tree.

We then present an algorithm that takes $O(k^3 \log N)$ time for updates and $O(k^2 \log N)$ time for queries in any causal tree. This can of course represent a tremendous speedup, especially for large networks. Our algorithm begins with a polynomial time *preprocessing* step, constructing another data structure (which is not itself a probabilistic tree) that supports fast queries and updates. The techniques we use are motivated by earlier algorithms for dynamic arithmetic trees, and involve "caching" sufficient intermediate computations during the update phase so that querying is also relatively easy. We note, however, that there are substantial and interesting differences between the algorithm for probabilistic networks and those for arithmetic trees. In particular, as will be apparent later, computation in probabilistic trees requires both bottom-up and top-down processing, whereas arithmetic trees need only the former. Perhaps even more interesting is that the relevant probabilistic operations have a different algebraic structure than arithmetic operations (for instance, they lack distributivity).

Bayesian causal trees have many applications in the literature including classification, [DH73, RKSA94], computer vision [HOW92, Che90], signal processing [Wil93], game playing [DK92], and statistical mechanics [BY90]. Nevertheless, causal trees are fairly limited for modeling purposes. However similar structures, called *join trees*, arise in the course of one of the standard algorithms for computing with arbitrary Bayesian networks (see [LS88]). Thus our algorithm for causal trees has potential relevance to many networks that are not trees. Because join trees have some special structure, they allow some optimization of the basic causal-tree algorithm. We elaborate on this in Section 4.

In Section 5 we consider the case of arbitrary polytrees. Although we suspect that it is possible to directly state an $O(\log N)$ algorithm for updates and queries in polytrees, in a similar fashion to the causal-tree algorithm of Section 2, we propose a different technique in this paper. This involves transforming the polytree to a join tree, and then using the results of Sections 2 and 4. The join tree of a polytree has a particularly simple form, giving an algorithm in which updates take $O(k^{p+3} \log N)$ time and queries $O(k^{p+2} \log N)$, where $p$ is the maximum number of parents of any node. Although the constant appears large, it must be noted that the original polytree takes $O(k^{p+1})$ space merely to represent, if conditional probability tables are given as explicit matrices.

## 2   CAUSAL TREES

Following [Pea88] we assume that a probabilistic causal tree is a directed tree in which each node represents a discrete random variable $X$, and each directed edge represents a matrix of conditional probabilities $M_{Y|X}$ (associated with edge $X \to Y$). Without loss of generality, we can assume all trees are binary trees, and are complete (*i.e.*, each node has 0 or 2 children.)[1] In a causal tree, each node (with the exception of the root) has only one incoming directed edge. The leaf nodes in the tree are called **evidence nodes**. Each leaf node is instantiated to a boolean vector with a single 1 corresponding to the observed evidence. Without loss of generality we assume evidence nodes are always leaves in the causal tree. The probability distribution of any non-evidence node in the tree is a probability vector $Bel(X)$ defined by

$$Bel(X) = p(X| \text{all evidence values})$$

and describes the discrete probability distribution of $X$ conditioned by the current instantiation of evidence nodes. The belief vector of an evidence node $X$ is fully known and does not depend on other nodes in the tree. In this paper we assume that the size of each vector $Bel(X)$ is a constant $k$; that is, the variable $X$ takes exactly $k$ values. A common case is that of binary random variables ($k = 2$) that take the value TRUE with probability $p$ and FALSE with probability $1 - p$.

---
[1]Converting any tree into one of this form involves at most doubling the number of nodes.

The main algorithmic problem is to compute $Bel(X)$ for each non-evidence node $X$ in the tree given the current evidence. Let $X$ be a node in a tree, with $U$ a parent of $X$, and $Y$ and $Z$ children of $X$. Intuitively speaking, the independence assumption in a causal tree states that the conditional probability of a node $U$ given the probability of $X$, $Y$ and $Z$ depends only on the probability of $X$; and the probability of $Y$ given $X$, $Z$ and $U$ depends only on $X$. We refer the reader to [Pea88] for details related to the modeling of independence assumptions using graphs.

As a result, the probability vector $Bel(X)$ can be computed in linear time (in the size of the tree) by a popular algorithm that simply solves the following set of equations.

$$Bel(X) = p(X| \text{all evidence}) = \alpha * \lambda(X) * \pi(X)$$

where $\alpha$ is a normalizing constant, $\lambda(X)$ is the probability of all the evidence in the subtree below node $X$ given $X$, and $\pi(X)$ is the probability of $X$ given all evidence in the rest of the tree. Let $X = (x_1, x_2, \ldots, x_k)$ and $(Y = y_1, y_2, \ldots, y_k)$ be two vectors. We define $*$ to be the component-wise product (pairwise or dyadic product of vectors):

$$X * Y = (x_1 y_1, x_2 y_2, \ldots, x_k y_k)$$

We define $\lambda(X)$ and $\pi(X)$ recursively as follows:

1. If $X$ is the root node, $\pi(X)$ is the prior probability of $X$.

2. If $X$ is a leaf node, $\lambda(X) = E$, where $E$ is the given evidence probability vector, that is a a boolean vector where a single 1 in the $i - th$ position corresponds to the observed evidence.

3. Otherwise, if, as shown in Figure 1, the children of node $X$ are $Y$ and $Z$, its sibling is $V$ and its parent is $U$, we have:
$$\lambda(X) = (M_{Y|X} \cdot \lambda(Y)) * (M_{Z|X} \cdot \lambda(Z))$$
$$\pi(X) = M_{X|U}^{\mathrm{T}} \cdot (\pi(U) * (M_{V|U} \cdot \lambda(V)))$$

Our derivation of belief vectors in causal trees is borrowed from [Pea88]. However, we use a somewhat different notation in that we don't describe messages sent to parents or successors, but rather discuss the direct relations among the $\pi$ and $\lambda$ vectors in terms of simple algebraic equations. We will take advantage of algebraic properties of these equations in our development.

We assume that the size of the causal tree is $N$, and that each variable has a constant domain size $k$. It follows that the matrices of conditional probabilities have size $k^2$. It is very easy to see that the equations above can be evaluated in time proportional to the size of the network. The formal proof is given in [Pea88].

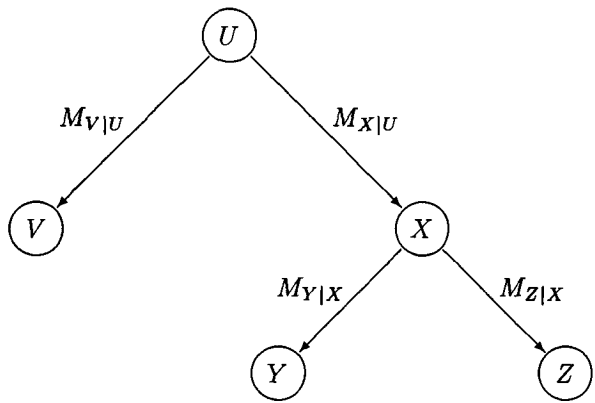

Figure 1: A segment of a causal tree.

**Theorem 1:** The belief distribution of every variable (that is, the marginal probability distribution for each variable, given the evidence) in a causal tree can be evaluated in $O(k^2 N)$ time where $N$ is the size of the tree. (The factor $k^2$ is due to the multiplication of a matrix by a vector that must be performed at each node.)

This theorem shows that it is possible to perform evidence absorption in $O(N)$ time, and queries in constant time (*i.e.*, by retrieving the previously computed values from a lookup table). In the next sections we will show how to perform both queries and updates in worst-case $O(\log N)$ time. Intuitively, we will not recompute all the marginal distributions after an update, but rather make only a small number of changes, sufficient, however, to compute the value of any variable with only a logarithmic delay.

## 2.1 A SIMPLE PREPROCESSING APPROACH

To obtain intuition about the new approach we begin with a very simple observation. Consider a causal tree $T$ of depth $D$. For each node $X$ in the tree we initially compute its $\lambda(X)$ vector. $\pi$ vectors are left un-computed. Given an update to a node $Y$, we calculate the revised $\lambda(X)$ vectors for all nodes $X$ that are ancestors of $Y$ in the tree. This clearly can be done in time proportional to the depth of the tree, $O(D)$. The rest of the information in the tree remains unchanged. Now consider a QUERY-NODE operation for some node $V$ in the tree. We obviously already have the accurate $\lambda(V)$ vector for every node in the tree including $V$. However, in order to compute its $\pi(V)$ vector we need to compute only the $\pi(Y)$ vectors for all the nodes above $V$ in the tree and multiply these by the appropriate $\lambda$ vectors that are kept current. This means that to compute the accurate $\pi(V)$ vector we need to perform $O(D)$ work as well. Thus, in this approach we don't perform the complete update to every $\lambda(X)$ and $\pi(X)$ vector in the tree.

**Lemma 2:** UPDATE-NODE and QUERY-NODE operations in a causal tree $T$ can be performed in $O(k^2 D)$ time where $D$ is the depth of the tree.

This implies that if the tree is balanced, both operations can be done in $O(\log N)$ time. However, in some important applications the trees are not balanced (e.g., models of temporal sequences [DKGH93]). The obvious question therefore is: Given a causal tree $T$ can we produce an equivalent balanced tree $T'$? While the answer to this question appears to be difficult, it is possible to a use a more sophisticated approach to produce a data structure (which is *not* a causal tree) to process queries and updates in $O(\log N)$ time. This approach is described in the subsequent sections.

## 2.2 A DYNAMIC DATA STRUCTURE FOR CAUSAL TREES

The data structure that will allow efficient incremental processing of a probabilistic tree $T = T_0$ will be a sequence of trees, $T_1, T_2, \ldots, T_i, \ldots, T_{\log N}$. Each $T_{i+1}$ will be a contracted version of $T_i$, containing half as many leaves as its predecessor.

For every $T_i$ there will be a set of equations for the $\pi$ and $\lambda$ values of nodes in $T_i$. In general, for each node $x$ in $T_i$ we will have equations of the form:[2]

$$\lambda(x) = A_i(x) \cdot \lambda(y) * B_i(x) \cdot \lambda(z)$$
$$\pi(x) = D_i(x) \cdot (\pi(u) * C_i(x) \cdot \lambda(v))$$

where $y$ and $z$ are the children of $x$, $v$ its sibling, and $u$ its parent in $T_i$. $A_i(x)$, $B_i(x)$, $C_i(x)$ and $D_i(x)$ are constant matrices. $C_i(x)$ is equal to either $A_i(u)$ or $B_i(u)$, and $D_i(x)$ is $B_i(u)^T$ or $A_i(u)^T$, depending on whether $x$ is the right or left child, respectively, of $u$. Note that it follows that we can identify $A_i(x), B_i(x), C_i(x), D_i(x)^T$ as the appropriate conditional probability matrices in the contracted tree, although we do not use this property in the following.

Below we list a few simple algebraic properties of the computations we perform during the contraction operation. We use $\text{Diag}_\alpha$ to denote the diagonal matrix whose diagonal entries are the components of the vector $\alpha$.

**Lemma 3:** Given (column) vectors $\alpha$, $\beta$ and $\gamma$, and matrix (of conditional probabilities) $M$ we have the following commutativity and associativity properties:

$$\alpha * \beta = \beta * \alpha$$
$$\alpha * (\beta * \gamma) = (\alpha * \beta) * \gamma$$
$$M \cdot (\alpha * \beta) = (M \cdot \text{Diag}_\alpha) \cdot \beta$$
$$(\alpha * \beta)^T \cdot M = \alpha^T \cdot (\text{Diag}_\beta \cdot M)$$

---

[2]Throughout, we assume that $*$ has lower precedence than conventional matrix multiplication (indicated by $\cdot$).

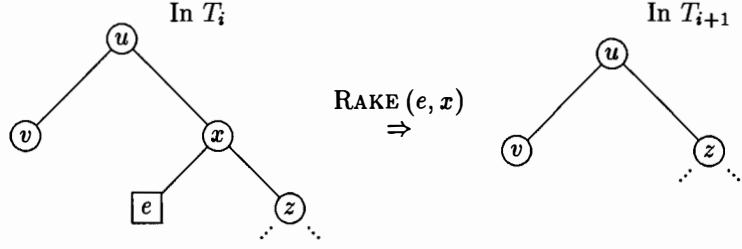

Figure 2: The effect of the operation RAKE $(e, x)$. $e$ must be a leaf, but $z$ may or may not be a leaf.

## 2.3 RAKE OPERATION

The basic operation used to contract the tree is RAKE which removes both a leaf and its parent from the tree. The effect of this operation on the tree is shown in Figure 2.

We now define the algebraic effect of this operation on the equations associated with the tree above. Intuitively, this operation simply substitutes the equations for $\lambda(x)$ and $\pi(x)$ into the equations for $\lambda(u)$, $\pi(z)$, and $\pi(v)$ (note that $\pi(u)$, $\lambda(z)$ and $\lambda(v)$ are unaffected). After having done this, nodes $x$ and $e$ and their associated equations can be removed from the tree.

Using the properties given in the Lemma above, for $\lambda(u)$ we obtain:

$$\begin{aligned}\lambda(u) &= A_i(u) \cdot \lambda(v) * B_i(u) \cdot \lambda(x) \\ &= A_i(u) \cdot \lambda(v) * B_i(u) \\ &\quad \cdot (A_i(x) \cdot \lambda(e) * B_i(x) \cdot \lambda(z)) \\ &= A_i(u) \cdot \lambda(v) * B_i(u) \\ &\quad \cdot \left(\text{Diag}_{A_i(x) \cdot \lambda(e)} \cdot B_i(x) \cdot \lambda(z)\right) \\ &= A_i(u) \cdot \lambda(v) * \\ &\quad \left(B_i(u) \cdot \text{Diag}_{A_i(x) \cdot \lambda(e)} \cdot B_i(x)\right) \cdot \lambda(z) \\ &= A_{i+1}(u) \cdot \lambda(v) * B_{i+1}(u) \cdot \lambda(z)\end{aligned}$$

where $A_{i+1}(u) = A_i(u)$ and
$B_{i+1}(u) = B_i(u) \cdot \text{Diag}_{A_i(x) \cdot \lambda(e)} \cdot B_i(x)$.

Similarly, for $\pi(z)$ we will have:

$$\begin{aligned}\pi(z) &= D_i(z) \cdot (\pi(x) * C_i(z) \cdot \lambda(e)) \\ &= D_{i+1}(z) \cdot (\pi(u) * C_{i+1}(z) \cdot \lambda(v))\end{aligned}$$

where $C_{i+1}(z) = C_i(x)$ and
$D_{i+1}(z) = D_i(z) \cdot \text{Diag}_{C_i(z) \cdot \lambda(e)} \cdot D_i(x)$.

For $\pi(v)$ we will have:

$$\begin{aligned}\pi(v) &= D_i(v) \cdot (\pi(u) * C_i(v) \cdot \lambda(x)) \\ &= D_{i+1}(v) \cdot (\pi(u) * C_{i+1}(v) \cdot \lambda(z))\end{aligned}$$

where $D_{i+1}(v) = D_i(v)$ and
$C_{i+1}(v) = C_i(v) \cdot \text{Diag}_{A_i(x) \cdot \lambda(e)} \cdot B_i(x) = B_{i+1}(u)$.

Note that since node $e$ is a leaf, the value of $\lambda(e)$ is known. The situation is entirely symmetric if the leaf being raked is a right child.

Beginning with the given tree $T = T_0$, each successive tree is constructed by performing a sequence of rakes, so as to rake away about half of the remaining evidence nodes. More specifically, let CONTRACT be the operation in which we apply the RAKE operation to every other leaf of a causal tree, in left-to-right order, excluding the leftmost and the rightmost leaf. Let $\{T_i\}$ be the set of causal trees constructed so that $T_{i+1}$ is the causal tree generated from $T_i$ by a single application of CONTRACT. The following result is proved using an easy inductive argument:

**Theorem 4:** Let $T_0$ be a causal tree of size $N$. Then the number of leaves in $T_{i+1}$ is equal to half the leaves in $T_i$ (not counting the two extreme leaves) so that starting with $T_0$, after $O(\log N)$ applications of CONTRACT, we produce a three-node tree: the root, the leftmost leaf and the rightmost leaf.

Below are a few observations about this process:

1. The complexity of CONTRACT is linear in the size of the tree. Additionally, $\log N$ applications of CONTRACT reduce the set of tree equations to a single equation involving the root in $O(N)$ total time.

2. The total space to store all the sets of equations associated with $\{T_i\}_{0 \leq i \leq \log N}$ is about twice the space required to store the equations for $T_0$.

3. With each equation in $T_{i+1}$ we also store equations that describe the relationship between the new coefficients (matrices) in $T_{i+1}$ to the matrices in $T_i$. We regard these equations as part of $T_{i+1}$. So, formally speaking $\{T_i\}$ are not pure causal trees but causal trees augmented with some auxiliary equations.

We note that the ideas behind the RAKE operation were originally developed by Miller-Reif [MR85] in the context of parallel computation of bottom-up arithmetic expression trees [KD88, KR90]. In contrast, we are using it in the context of incremental update and query operations in sequential computing. A similar data structure to ours was independently proposed in [Fre93] in the context of dynamic arithmetic expression trees, and a different approach for incremental computing on arithmetic trees was developed in [CT91]. There are important and interesting differences between the arithmetic expression tree case and our own. For arithmetic expressions all computation is done bottom-up. However, in probabilistic networks $\pi$-messages must be passed top-down. Furthermore, in arithmetic expressions when two algebraic operations are allowed, we typically require the distributivity of one operation over the other, but the analogous property does not hold for us. In these respects our approach is a substantial generalization of the previous work, while remaining conceptually simple and practical.

## 2.4 PERFORMING QUERIES AND UPDATES EFFICIENTLY

In this section we shall show how to utilize the contracted trees $T_i$, $0 \leq i \leq \log N$ to perform queries and updates in $O(\log N)$ time in general causal trees. We shall show that a logarithmic amount of work will be necessary and sufficient to compute enough information in our data structure to update and query any $\lambda$ or $\pi$ value. Section 3 provides a simple illustration of our method for a chain network.

## 2.5 $\lambda$ QUERIES

To compute $\lambda(x)$ for some node $x$ we can do the following. We first locate $ind(x)$, which is defined to be the highest level $i$ for which $x$ has equations. The equation for $\lambda(x)$ is of the form:

$$\lambda(x) = A_i(x) \cdot \lambda(y) * B_i(x) \cdot \lambda(z)$$

where $y$ and $z$ are the left and right children, respectively, of $x$ in $T_i$.

Since $x$ does not appear in $T_{i+1}$, it was raked at this level of equations, which implies that one child (we assume $z$) is a leaf. We therefore only need to compute $\lambda(y)$, which can be done recursively. If instead $y$ was the raked leaf, we would compute $\lambda(z)$ recursively.

In either case $O(1)$ operations are done in addition to one recursive call, which is to a value at a higher level of equations. Since there are $O(\log N)$ levels, and the only operations are matrix by vector multiplications, the procedure takes $O(k^2 \log N)$ time. The function $\lambda$-QUERY $(x)$ is given in Figure 3.

## 2.6 UPDATES

We now describe how the update operations can modify enough information in the data structure to allow us to query the $\lambda$ vectors and $\pi$ vectors efficiently. Most importantly the reader should note that the update operation does not try to maintain the correct $\pi$

---

FUNCTION $\lambda$-QUERY $(x)$

We look up the equation associated with $\lambda(x)$ in $T_{ind(x)}$.

Case 1: $x$ is a leaf. Then the equation is of the form: $\lambda(x) = e$ where $e$ is known. In this case we return $e$.

Case 2: The equation associated with $\lambda(x)$ is of the form

$$\lambda(x) = A_i(x) \cdot \lambda(y) * B_i(x) \cdot \lambda(z)$$

where $z$ is a leaf and therefore $\lambda(z)$ is known. In this case we return

$$A_i(X) \cdot \lambda\text{-QUERY}(y) * B_i(X) \cdot \lambda(z)$$

The case where $y$ is the leaf is analogous.

Figure 3: Function to compute the $\lambda$ value of a node.

---

FUNCTION UPDATE $(Term = Value, i)$

1. Find the (at most one) equation in $T_i$, defining some $A_i$ or $B_i$, in which Term appears on the right-hand side; let $Term'$ be the matrix defined by this equation (i.e., its left-hand side).

2. Update $Term'$; let Value be the new value.

3. Call UPDATE $(Term' = Value, i + 1)$ recursively.

Figure 4: The update procedure.

---

and $\lambda$ values. It is sufficient to ensure that, for all $i$ and $x$, the matrices $A_i(x)$ and $B_i(x)$ (and thus also $C_i(x)$ and $D_i(x)$) are always up to date.

When we update the value of an evidence node, we are simply changing the $\lambda$ value of some leaf $e$. At each level of equations, the value of $\lambda(e)$ can appear at most twice: once in the $\lambda$-equation of $e$'s parent and once in the $\pi$-equation of $e$'s sibling in $T_i$. When $e$ disappears, say at level $i$, its value is incorporated into one of the constant matrices $A_{i+1}(u)$ or $B_{i+1}(u)$ where $u$ is the grandparent of $e$ in $T_i$. This constant matrix in turn affects exactly one constant matrix in the next higher level, and so on. Since the effect at each level can be computed in $O(k^3)$ time (due to matrix multiplication) and there are $O(\log N)$ levels of equations, the update can be accomplished in $O(k^3 \log N)$ time. The constant $k^3$ is actually pessimistic, because faster matrix multiplication algorithms exist.

The update procedure is given in Figure 4. UPDATE is initially called as UPDATE$(\lambda(E) = e, i)$ where $E$ is a leaf, $i$ the level at which it was raked, and $e$ is the new evidence. This operation will start a sequence of $O(\log N)$ calls to function UPDATE $(X = \text{Term}, i)$ as the change will propagate to $\log N$ equations.

## 2.7  $\pi$ QUERIES

It is relatively easy to use a similar recursive procedure to perform $\pi(x)$ queries. Unfortunately, this approach yields an $O(\log^2 N)$-time algorithm if we simply use recursion to calculate $\pi$ terms and calculate $\lambda$ terms using our earlier procedure. This is because there will be $O(\log N)$ recursive calls to calculate $\pi$ values, but each is defined by an equation that also involves $\lambda$ term taking $O(\log N)$ time to compute.

To achieve $O(\log N)$ time, we shall instead implement $\pi(x)$ queries by defining a procedure CALC$\pi\lambda(x,i)$ which returns a triple of vectors $\langle P, L, R\rangle$ such that $P = \pi(x)$, $L = \lambda(y)$ and $R = \lambda(z)$ where $y$ and $z$ are the left and right children, respectively, of $x$ in $T_i$.

To compute $\pi(x)$ for some node $x$ we can do the following. We first find the highest level $i$ for which $x$ has equations. The equation for $\pi(x)$ is of the form:

$$\pi(x) = D_i(x) \cdot (\pi(u) * C_i(x) \cdot \lambda(v))$$

where $u$ is the parent of $x$ in $T_i$ and $v$ its sibling. We then call procedure CALC$\pi\lambda(u, i+1)$ which will return the triple $\langle \pi(u), \lambda(v), \lambda(x)\rangle$, from which we immediately can compute $\pi(x)$ using the above equation.

Procedure CALC$\pi\lambda(x,i)$ can be implemented in the following fashion.

Case 1: If $T_i$ is a 3-node tree with $x$ as its root, then both children of $x$ are leaves, hence their $\lambda$ values are known, and $\pi(x)$ is a given sequence of prior probabilities for $x$.

Case 2: If $x$ does not appear in $T_{i+1}$, then one of $x$'s children is a leaf, say $e$ which is raked at level $i$. Let $z$ be the other child. We call CALC$\pi\lambda(u, i+1)$, where $u$ is the parent of $x$ in $T_i$, and receive back $\langle \pi(u), \lambda(z), \lambda(v)\rangle$. We can now compute $\pi(x)$ from $\pi(u)$ and $\lambda(v)$, and we have $\lambda(e)$ and $\lambda(z)$, so we can return the necessary triple.
Specifically,

$$\pi(x) = \begin{cases} D_i(x) \cdot (\pi(u) * A_{i+1}(u) \cdot \lambda(v)) \\ D_i(x) \cdot (\pi(u) * B_{i+1}(u) \cdot \lambda(v)) \end{cases}$$

where the choice depends on whether $x$ is the right or left child, respectively, of $u$ in $T_i$.

Case 3: If $x$ does appear in $T_{i+1}$, then we call CALC$\pi\lambda(x, i+1)$. This returns the correct value of $\pi(x)$. For any child $z$ of $x$ in $T_i$ that remains a child of $x$ in $T_{i+1}$, it also returns the correct value of $\lambda(z)$. If $z$ is a child of $x$ that does not occur in $T_{i+1}$, then it must be the case that $z$ was raked at level $i$ so that one of $z$'s children, say $e$, is a leaf and let the other child be $q$. In this situation CALC$\pi\lambda(x, i+1)$ has returned the value of $\lambda(q)$ and we can compute

$$\lambda(z) = A_i(z) \cdot \lambda(e) * B_i(z) \cdot \lambda(q)$$

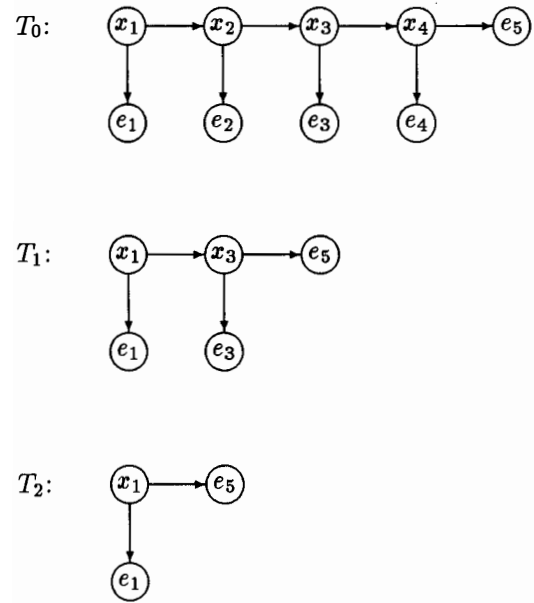

Figure 5: A simple chain example.

and return this value.

In all three cases, there is a constant amount of work done in addition to a single recursive call that uses equations at a higher level. Since there are $O(\log N)$ levels of equations, each requiring only matrix by vector multiplication, the total work done is $O(k^2 \log N)$.

## 3   EXAMPLE: A CHAIN

To obtain an intuition about the algorithms, we sketch how to generate and utilize the $T_i$, $0 \leq i \leq \log N$ and their equations to perform $\lambda$-value queries and updates in $O(\log N)$ time on an $N = 2L+1$ node chain of length $L$. Consider the chain of length 4 in Figure 5, and the trees that are generated by repeated application of CONTRACT to the chain.

The equations that correspond to the contracted trees in the figure are as follows (ignoring trivial equations):

$$\left.\begin{aligned}\lambda(x_1) &= A_0(x_1) \cdot \lambda(e_1) * B_0(x_1) \cdot \lambda(x_2) \\ \lambda(x_2) &= A_0(x_2) \cdot \lambda(e_2) * B_0(x_2) \cdot \lambda(x_3) \\ \lambda(x_3) &= A_0(x_3) \cdot \lambda(e_3) * B_0(x_3) \cdot \lambda(x_4) \\ \lambda(x_4) &= A_0(x_4) \cdot \lambda(e_4) * B_0(x_4) \cdot \lambda(e_5)\end{aligned}\right\} \text{ for } T_0$$

$$\left.\begin{aligned}\lambda(x_1) &= A_1(x_1) \cdot \lambda(e_1) * B_1(x_1) \cdot \lambda(x_2) \\ \lambda(x_3) &= A_1(x_3) \cdot \lambda(e_3) * B_1(x_3) \cdot \lambda(e_5) \\ \text{where}& \\ B_1(x_1) &= B_0(x_1) \cdot \text{Diag}_{A_0(x_2) \cdot \lambda(e_2)} \cdot B_0(x_2) \\ B_1(x_3) &= B_0(x_3) \cdot \text{Diag}_{A_0(x_4) \cdot \lambda(e_4)} \cdot B_0(x_4)\end{aligned}\right\} \text{ for } T_1$$

$$\left.\begin{array}{l}\lambda(x_1) = A_2(x_1) \cdot \lambda(e_1) * B_2(x_1) \cdot \lambda(e_5) \\ \text{where} \\ B_2(x_1) = B_1(x_1) \cdot \text{Diag}_{A_1(x_3) \cdot \lambda(e_3)} \cdot B_1(x_3)\end{array}\right\} \text{ for } T_2$$

Now consider a query operation on $x_2$. Rather than performing the standard computation we will find the level where $x_2$ was "raked". This gives us the equation

$$\lambda(x_2) = A_0(x_2) \cdot \lambda(e_2) * B_0(x_2) \cdot \lambda(x_3)$$

Now we find where $x_3$ is "raked". That happened on level 1. However, on that level the equation associated with $x_3$ is:

$$\lambda(x_3) = A_1(x_3) \cdot \lambda(e_3) * B_1(x_3) \cdot \lambda(e_5)$$

That means that we need not follow down the chain. In general for a chain of $N$ nodes we can answer any query to a node on the chain by evaluating $\log N$ equations instead of $N$ equations.

Now consider an update for $e_4$. Since $e_4$ was raked immediately, we first modify the equation

$$B_1(x_3) = B_0(x_3) \cdot \text{Diag}_{A_0(x_4) \cdot \lambda(e_4)} \cdot B_0(x_4)$$

on the first level where $e_4$ occurs on the right-hand side. Since $B_1(x_3)$ is affected by the change to $e_4$, we subsequently modify the equation

$$B_2(x_1) = B_1(x_1) \cdot \text{Diag}_{A_1(x_3) \cdot \lambda(e_3)} \cdot B_1(x_3)$$

on the second level.

## 4  JOIN TREES

Perhaps the best-known technique for computing with arbitrary (*i.e.*, *not* singly-connected) Bayesian networks uses the idea of *join trees* (junction trees) [LS88]. In many ways a join tree can be thought of as a causal tree, albeit one with somewhat special structure. Thus the algorithm in the previous section can be applied. However, the structure of a join tree permits some optimization, which we describe in this section. This becomes especially relevant in the next section, where we use the join-tree technique to show how $O(\log N)$ updates and queries can be done for arbitrary polytrees. Our review of join-trees and their utility is necessarily extremely brief and quite incomplete; for clear expositions see, for instance, [SDLC93, Pea88].

Given any Bayesian network, the first step towards constructing a join-tree is to *moralize* the network: insert edges between every pair of parents of a common node, and then treat all edges in the graph as being undirected [SDLC93]. The resulting undirected graph is called the *moral* graph. We are interested in undirected graphs that are *chordal*: every cycle of length 4 or more should contain a chord (*i.e.*, an edge between two nodes that are non-adjacent in the cycle). If the moral graph is not chordal, it is necessary to add edges to make it so; various techniques for this *triangulation* stage are known (for instance, see [SDLC93]).

The graph constructed this way has many useful properties. If $p$ is a probability distribution represented in a Bayesian network $G = (V, E)$, and $M = (V, F)$ is the result of moralizing and then triangulating $G$, then:

1. $M$ has at most $|V|$ maximal cliques, say, $C_1, \ldots, C_{|V|}$.
2. The cliques can be ordered so that for each $i > 1$ there is a unique $j(i) < i$ such that
$$C_i \cap C_{j(i)} = C_i \cap (C_1 \cup C_2 \cup \ldots \cup C_{i-1}.)$$
The tree $T$ formed by treating the cliques as nodes, and connecting each node $C_i$ to its "parent" $C_{j(i)}$, is called a *join tree*.
3. $p = \prod_{i=1}^{k} p(C_i | C_{j(i)})$
4. $p(C_i | C_{j(i)}) = p(C_i | C_{j(i)} \cap C_i)$

From 2 and 3, we see that if we direct the edges in $T$ away from the "parent" cliques, the resulting directed tree is in fact a Bayesian causal tree that can represent the original distribution $p$. This is true no matter what the form of the original graph. Of course, the price is that the cliques may be large, and so the domain size (the number of possible values of a clique node) can be of exponential size. This is why this technique is not guaranteed to be efficient.

We can use the RAKE technique of Section 2 on the directed join tree without any modification. However, property 4 above shows that the conditional probability matrices in the join tree have a special structure. We can use this to gain some efficiency. In the following, let $k$ be the domain size of the variables in $G$ as usual. Let $n$ be the maximum size of cliques in the join tree; without loss of generality we can assume that all cliques are of the same size (by adding "dummy" variables if necessary). Thus the domain size of each clique is $K = k^n$. Finally, let $c$ be the maximum intersection size of a clique and its parent (*i.e.* $|C_{j(i)} \cap C_i|$) and $L = k^c$.

In the standard algorithm, we would represent $p(C_i | C_{j(i)})$ as a $K \times K$ matrix, $M_{C_i | C_{j(i)}}$. However, $p(C_i | C_{j(i)} \cap C_i)$ can be represented as a smaller $L \times K$ matrix, $M_{C_i | C_{j(i)} \cap C_i}$. By property 4 above, $M_{C_i | C_{j(i)}}$ is identical to $M_{C_i | C_{j(i)} \cap C_i}$, except that many rows are repeated. Thus there is a $K \times L$ matrix $J$ such that

$$M_{C_i | C_{j(i)}} = J \cdot M_{C_i | C_{j(i)} \cap C_i}.$$

($J$ is actually a simple matrix whose entries are 0 and 1, with exactly one 1 per row; however we shall not need to use this fact.)

Our claim is that, in the case of join trees, the following is true. First, all the matrices $A_i, B_i, C_i, D_i$ used in the RAKE algorithm can be stored in factored form, as the product of two matrices of dimension $K \times L$ and $L \times K$ respectively. So, for instance, we factor $A_i$ as $A_i^l \cdot A_i^r$. We never need to explicitly compute, or store, the full matrices. As we have just seen, this claim is true when $i = 0$ because the $M$ matrices factor this way. The proof for $i > 1$ uses an inductive argument, which we illustrate below. The second claim is that, when the matrices are stored in factored form, all the matrix multiplications used in the RAKE algorithm are of one of the following types: 1) an $L \times K$ matrix times a $K \times L$ matrix, 2) an $L \times K$ matrix times a $K \times K$ diagonal matrix, 3) an $L \times L$ matrix times an $L \times K$ matrix, or 4) an $L \times K$ matrix times a vector.

To prove these claims consider, for instance, the equation defining $B_{i+1}$ in terms of lower-level matrices. From Section 2, $B_{i+1}(u) = B_i(u) \cdot \text{Diag}_{A_i(x) \cdot \lambda(e)} \cdot B_i(x)$. But, by assumption, this is:

$(B_i^l(u) \cdot B_i^r(u)) \cdot \text{Diag}_{(A_i^l(x) \cdot A_i^r(x)) \cdot \lambda(e)} \cdot (B_i^l(x) \cdot B_i^l(x))$,

which, using associativity, is clearly equivalent to

$B_i^l(u) \cdot (((B_i^r(u) \cdot \text{Diag}_{A_i^l(x) \cdot (A_i^r(x) \cdot \lambda(e))}) \cdot B_i^l(x)) \cdot B_i^l(x))$.

However, every multiplication in this expression is one of the forms stated earlier. Identifying $B_{i+1}^l(u)$ as $B_i^l(u)$ and $B_{i+1}^r(u)$ as the remainder proves this case. The equations for the matrices $A_i, C_i, D_i$ are similar. Thus, even using the most straightforward technique for matrix multiplication, the cost of updating $B_{i+1}$ is $O(KL^2) = O(k^{n+2c})$. This contrasts with $O(K^3)$ if we do not factor the matrices, and may represent a worthwhile speedup if $c$ is small. Note that the overall time for an update using this scheme is $O(k^{n+2c} \log N)$. Queries, which only involve matrix by vector multiplication, require $O(k^{n+c} \log N)$ time.

For many join trees the difference between $N$ and $\log N$ is unimportant, because the clique domain size $K$ is often enormous and dominates the complexity. Of course, in any such case our technique has little to offer. But in other cases the benefits will be worthwhile. The most important general class in which this is so, and our immediate reason for presenting the technique for join trees, is the case of polytrees.

## 5  POLYTREES

A polytree is a singly connected Bayesian network; we drop the assumption of Section 2 that each node has at most one parent. Polytrees offer much more flexibility than causal trees, and yet there is a well-known process that can update and query in $O(N)$ time, just as for causal trees. For this reason polytrees are an extremely popular class of networks.

We suspect that it is possible to present an $O(\log N)$ algorithm for updates and queries in polytrees as a direct extension of the ideas in Section 2. Instead we propose a different technique, which involves converting a polytree to its join tree and then using the ideas of the preceding section. The basis for this is the simple observation that the join tree of a polytree is already chordal. Thus (as we show in detail below) little is lost by considering the join tree instead of the original polytree. The specific property of polytrees that we require is the following. We omit the proof.

**Proposition 5:** If $T$ is the moral graph of a polytree $P = (V, E)$ then $T$ is chordal, and the set of maximal cliques in $T$ is $\{\{v\} \cup parents(v) : v \in V\}$.

Let $p$ be the maximum number of parents of any node. From the proposition, every maximal clique in the join tree has at most $p + 1$ variables, and so the domain size of a node in the join tree is $K = k^{p+1}$. This may be large, but recall that the conditional probability matrix in the original polytree, for a variable with $p$ parents, has $K$ entries anyway. Thus $K$ is really a measure of the size of the polytree itself.

It now follows from the proposition above that we can perform query and update in polytrees in time $O(K^3 \log N)$, simply by using the algorithm of Section 2 on the directed join tree. But, as noted in Section 4, we can do better. Recall that the savings depend on $c$, the maximum size of the intersection between any node and its parent in the join tree. However, when the join tree is formed from a polytree, no two cliques can share more than a single node. This follows immediately from Proposition 5, for if two cliques have more than one node in common then there must be either two nodes that share more than one parent, or else a node and one of its parents that both share yet another parent. Neither of these is consistent with the network being a polytree. Thus in the complexity bounds of Section 4, we can put $c = 1$. It follows that we can process updates in $O(Kk^{2c} \log N) = O(k^{p+3} \log N)$ time and queries in $O(k^{p+2} \log N)$.

## 6  CONCLUSION

We conclude with a few additional perspectives on these results. One interesting issue is the potential for a parallel implementation. The standard polytree algorithm is often presented in a message-passing framework, which in principle suggests some ready parallelism. Our approach also admits parallelism easily, but in a different way, namely, most of the work is in matrix multiplications. Parallel matrix operations are exceptionally efficiently supported on almost all modern parallel machines. Approaches based on matrix operations, such as the proposal in this paper, may offer practical advantages over other schemes. Clearly, this is a topic that calls for empirical investigation.

One experiment, although not specifically concerning parallelism, is reported in [DKGH93]. There we report an application with a probabilistic causal tree of depth 300 that consists of 600 nodes and all the matrices of conditional probabilities are $2 \times 2$. The tree is used to model the dependence of a protein's secondary structure on its chemical structure. The detailed description of the problem and experimental results are given in [DKGH93]. For this problem we can obtain an effective speed-up of about a factor of 10 to perform a query/update cycle as compared to the standard algorithm. Clearly, getting an order of magnitude improve-ment in response time could be of tremendous importance in future use of such systems. There are other intriguing potential applications of our approach in the biological domain, such as simulated muta-genesis; see [DKGH93] for discussion and references.

*Acknowledgements*

We thank Rao Kosaraju and Steven Omohundro for many helpful comments and Mike Goodrich for pointing out the relevant work by Greg Frederickson and Cohen & Tamassia. Part of this was done at NEC Research Institute, NJ, USA, during Simon Kasif's 1994-1995 sabbatical. Kasif's research at Johns Hopkins University has been supported in part by the National Science foundation under Grant No. IRI-9220960.